\ifdefined\pdfminorversion\pdfminorversion=7\fi
\documentclass{article}
\usepackage{leaplab}
\usepackage[table,x11names]{xcolor}

\usepackage[utf8]{inputenc} 
\usepackage[T1]{fontenc}    
\usepackage{url}            
\usepackage{booktabs}       
\usepackage{amsfonts}       
\usepackage{nicefrac}       
\usepackage{microtype}      
\usepackage{tabularx}   
\usepackage{enumitem}

\usepackage[utf8]{inputenc} 
\usepackage[T1]{fontenc}    
\usepackage{url}            
\usepackage{booktabs}       
\usepackage{amsfonts}       
\usepackage{nicefrac}       
\usepackage{microtype}      
\definecolor{citecolor}{HTML}{0071bc}
\definecolor{shadecolor}{HTML}{EFEFEF}
\definecolor{linkcolor}{HTML}{9A4D92}
\usepackage[colorlinks, linkcolor=linkcolor, colorlinks, anchorcolor=blue, citecolor=citecolor, pagebackref=True]{hyperref}

\usepackage{amsmath}
\usepackage{microtype}
\usepackage{graphicx}
\usepackage{booktabs} 
\usepackage{caption}
\usepackage{wrapfig}
\usepackage{float}

\usepackage{titletoc} 

\usepackage{subcaption}
\usepackage[export]{adjustbox}
\usepackage{footnote}
\usepackage{tablefootnote}
\usepackage{threeparttable}
\usepackage[capitalize,noabbrev]{cleveref}

\usepackage{tabularray}
\UseTblrLibrary{booktabs}
\usepackage{multirow}
\usepackage{multicol}
\usepackage{xspace}
\usepackage{mathrsfs}
\usepackage{pifont}
\usepackage{makecell}
\usepackage{CJK}

\usepackage{marvosym} 
\usepackage[most]{tcolorbox}

\definecolor{brand}{HTML}{000000}
\definecolor{xulabwine}{HTML}{5E3545}
\definecolor{titlebg}{HTML}{F7F2F3}

\hypersetup{citecolor=citecolor,linkcolor=linkcolor,filecolor=linkcolor,urlcolor=magenta}

\usepackage{titletoc}

\newenvironment{leapabstract}{\begin{tcolorbox}[
colback=titlebg,colframe=white,boxrule=0pt,arc=10pt,
left=16pt,right=16pt,top=12pt,bottom=12pt,width=\textwidth,
enlarge left by=0mm,before skip=10pt,after skip=10pt]}{\end{tcolorbox}}

\fancypagestyle{fancytitlepage}{\fancyhead{}\lhead{\color{brand}\Large\bfseries Xu Lab}\rhead{\color{brand}\itshape HKUST}\cfoot{}}

\usepackage{amssymb}
\input{halo-layout}

\newlength{\halocolumnwidth}
\newcommand{\method}{HALO}

\crefname{table}{Tab.}{Tabs.}
\Crefname{table}{Tab.}{Tabs.}
\crefname{figure}{Fig.}{Figs.}
\Crefname{figure}{Fig.}{Figs.}

\icmltitlerunning{HALO: Human-Aligned Data Collection for Mobile Manipulation}
\hypersetup{pdfauthor={Author One, Author Two, Author Three, Author Four, Author Five, Author Six}}

\begin{document}
\raggedbottom
\ifdefined\XeTeXversion\fi

\icmltitle{HALO: \textbf{H}uman-\textbf{A}ligned Data Collection for
\texorpdfstring{\\}{ }Who\textbf{L}e-body M\textbf{O}bile Manipulation}

\begin{icmlauthorlist}
\raggedright
\mbox{Shuai Yang$^{1,2*}$}, \mbox{Fangyi Xu$^{2,*}$}, \mbox{Yishu Shen$^{1}$},
\mbox{Kecheng Zheng$^{2}$} and \mbox{Yinghao Xu$^{1,\dagger }$}
\par
\end{icmlauthorlist}

\medskip
{\small
$^{1}$The Hong Kong University of Science and Technology\par
$^{2}$Robbyant\par
$^{*}$Equal Contribution. $^{\dagger}$Corresponding author (e-mail:\texttt{justimyhxu@ust.hk})\par
}
\vskip .3in

\begin{leapabstract}
Learning whole-body mobile manipulation requires demonstrations that coordinate bimanual actions with locomotion across extended workspaces.
However, existing interfaces trade off direct robot control, perception and action alignment, and natural operator mobility,
making such demonstrations difficult to collect.
We introduce \method{}, a human-aligned data collection system for whole-body mobile manipulation.
The system aligns the operator's perception and action space with the robot while preserving direct physical feedback and natural locomotion.
Operators guide the robot end-effectors through kinesthetic teaching and move a low-resistance platform without the deployed mobile base while performing bimanual manipulation.
This interface enables more natural whole-body demonstrations, but the close human--robot coupling introduces discrepancies between the demonstration and deployment settings.
We address these discrepancies with deployment-aligned demonstration processing, which calibrates and temporally aligns platform motion with the deployed base response and masks and inpaints demonstrator hands in recorded observations.
Compared with Mobile ALOHA and VR teleoperation, \method{} enables up to $2.9\times$ faster demonstration collection while maintaining comparable or higher success rates.
Policies trained with \method{} perform precise manipulation tasks, including zip-tie fastening,
and achieve a 73.3\% success rate on a six-stage long-horizon mobile manipulation task involving repeated coordinated base and arm motion. \href{https://robbyant-research.github.io/Lingbot-ALOHA/}{Project page}

\end{leapabstract}

\section{Introduction}

End-to-end robot policies and imitation learning~\cite{aloha,dp,kim2024openvla,black2024pi_0,va,vla} have achieved strong performance across a wide range of real-world manipulation tasks. However, most prior work focuses on tabletop settings, where robots are confined to a fixed workspace. Real-world applications instead require robots to extend their workspace through mobility while continuing to manipulate objects, motivating coordinated whole-body bimanual manipulation~\cite{mobile_aloha}. Learning such behaviors requires efficient collection of demonstrations that jointly capture locomotion and manipulation.

Existing approaches to whole-body data collection make different trade-offs in preserving natural human coordination.
Indirect teleoperation methods (e.g., VR-based interfaces~\cite{zhu2022bottom,rakita2018autonomous,honerkamp2025whole}) provide less direct manipulation control and lack force feedback.
Leader--follower systems (e.g., Mobile ALOHA~\cite{mobile_aloha}) provide more direct manipulation through leader arms and kinesthetic teaching, but the spatial separation between the operator's hands, viewpoint, and robot end-effectors introduces a perception gap and constrains natural operator mobility.

\halotopbreak
\halowraplabels{fig:motivation}
\begin{wrapfigure}{R}{\halocolumnwidth}
    \raggedleft
    \captionsetup{justification=raggedright,singlelinecheck=false}
    \includegraphics[width=1\linewidth]{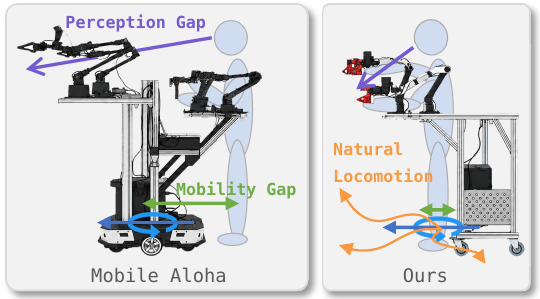}
    \caption{\textbf{Interface comparison.} Mobile ALOHA exhibits perception and mobility gaps, while our interface aligns human--robot perception and preserves natural operator mobility.}
    \label{fig:motivation}
\end{wrapfigure}

Retargeting-based approaches (e.g., UMI~\cite{umi} and HoMMI~\cite{xu2026hommi}) instead preserve unconstrained human motion by mapping human demonstrations to robot behavior through retargeting or whole-body control.
Taken together, existing approaches balance direct robot control, perception--action alignment, and natural operator mobility in different ways, often emphasizing one property over another.
These trade-offs across perception, manipulation, and mobility are summarized in~\cref{tab:survey}.

Effective whole-body demonstration requires preserving the robot's native action constraints and direct physical feedback while \textbf{aligning the operator's perception--action space} with the robot and \textbf{preserving the operator's natural mobility}, as illustrated in~\cref{fig:motivation}. We adopt two design principles:
First, the manipulation interface should preserve direct robot control and feedback. We therefore retain the robot arms and use kinesthetic teaching~\cite{akgun2012trajectories}, keeping the operator close to the end-effectors to reduce the \textbf{perception gap} while preserving direct visual and force feedback.
Second, the mobility interface should preserve the operator's natural locomotion without requiring control of the mobile base. We therefore remove the mobile base during demonstration, reducing the \textbf{mobility gap} and allowing natural body motion for repositioning.

While these interfaces enable more natural whole-body demonstrations, the resulting human--robot coupling introduces visual and control distribution shifts between demonstration data and autonomous deployment~\cite{xu2026hommi}. We identify these discrepancies and introduce a deployment aligned post-processing pipeline that removes demonstrator-induced artifacts from both visual observations and control signals, aligning the collected demonstrations with policy execution.

Together, these design choices improve demonstration efficiency and whole-body coordination. Compared with Mobile ALOHA and VR teleoperation, \method{} enables up to $2.9\times$ faster demonstration collection while maintaining higher success rates. The resulting policies achieve sub-millimeter manipulation precision and a 73.3\% success rate on a six-stage long-horizon mobile manipulation task involving repeated coordinated base and arm motion. These results demonstrate that human-aligned data collection supports efficient, dexterous, and coordinated whole-body policy learning. Our contributions are threefold:

\WFclear
\begin{itemize} 
\item \textbf{Human-aligned data collection system.} We introduce a whole-body bimanual teaching system that aligns the operator's perception--action space with the robot while preserving natural mobility, enabling efficient collection of coordinated and dexterous mobile manipulation.

\item \textbf{Deployment-aligned data post-process.} We identify and mitigate demonstrator-induced artifacts in visual observations and control, enabling demonstrations collected under tight human--robot coupling to be used for autonomous policy learning. 

\item \textbf{Systematic real-world evaluation.} We evaluate the proposed system in terms of data-collection efficiency and demonstration success, and we further study the effects of different inference strategies as well as the achievable precision and efficiency of the learned policies.

\end{itemize}

\begin{figure*}[!htbp]
    \centering
    \includegraphics[width=1\linewidth]{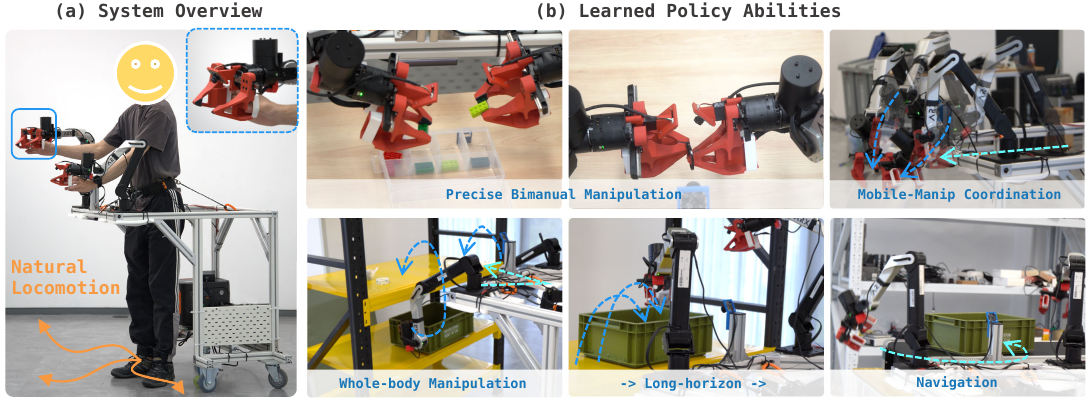}
    \caption{\textbf{\method{} overview and learned capabilities.} 
(a) Human-aligned data collection with natural locomotion and direct bimanual kinesthetic teaching. 
(b) Learned policies exhibit precise manipulation, coordinated whole-body control, long-horizon mobile manipulation, and navigation.}
    \label{fig:teaser}
\end{figure*}

\begin{table}[!htbp]
\centering
\caption{\textbf{Qualitative comparison of data collection systems.} We compare representative interfaces in terms of the desired properties for whole-body mobile manipulation data collection.}
\small
\setlength{\tabcolsep}{3pt}
\begin{tabularx}{\linewidth}{@{}>{\raggedright\arraybackslash}p{0.31\linewidth}*{6}{>{\centering\arraybackslash}X}@{}}
\toprule
Method 
& Mobile Support 
& Egocentric Perception 
& Force Feedback 
& Natural Locomotion 
& Retargeting-Free Actions
& Precise Base Control \\
\midrule
GELLO~\cite{gello} & $\times$   & $\times$   & $\times$   & $\times$   & \checkmark & $\times$ \\
Mobile ALOHA~\cite{mobile_aloha}           & \checkmark & $\times$   & $\times$   & $\times$   & \checkmark & \checkmark \\
VR~\cite{zhu2022bottom,rakita2018autonomous,honerkamp2025whole}                     & \checkmark & $\times$   & $\times$   & $\times$   & \checkmark   & \checkmark \\
Exoskeleton~\cite{citron2026modpack,liang2026universal}            & \checkmark & \checkmark   & \checkmark   & \checkmark   & $\times$   & $\times$ \\ 
UMI~\cite{umi} / HoMMI~\cite{xu2026hommi}                    & \checkmark &  \checkmark  & \checkmark & \checkmark & $\times$   & $\times$ \\ \midrule
\method{}              & \checkmark & \checkmark & \checkmark & \checkmark & \checkmark & \checkmark \\
\bottomrule
\end{tabularx}
\label{tab:survey}
\end{table}

\section{Related Work}

\subsection{Robot Demonstration Collection}




Traditional teleoperation minimizes the embodiment gap by directly controlling the target robot, but interfaces such as SpaceMouse devices, joysticks, and VR controllers remain indirect~\cite{zhu2022bottom,rakita2018autonomous,honerkamp2025whole,wong2022error}. Kinematic replicas, such as ALOHA~\cite{aloha,mobile_aloha}, GELLO~\cite{gello,jiang2025behavior}, and exoskeleton-based approaches~\cite{citron2026modpack,chen2026supersuit}, provide more intuitive control but typically lack force feedback, although some recent systems incorporate it~\cite{liang2026universal}; they also become cumbersome in mobile settings. UMI~\cite{umi} enables scalable in-the-wild data collection, while recent extensions~\cite{shen2026fomi,xu2026hommi,yu2025egomi} incorporate force feedback, egocentric sensing and whole-body controllers for mobile manipulation; however, base motion remains difficult to control precisely, and visual and control gaps persist. Our system instead aligns the operator with the robot, providing direct visual and force feedback.

\subsection{Human-to-Robot Demonstration Transfer}

Transferring human demonstrations to robot policy learning requires bridging the visual and action gaps between human embodiment and robot execution. DexUMI~\cite{xu2025dexumi} removes the human hand through inpainting and uses action replay to obtain the corresponding foreground dexterous robot hand. 

\halotopbreak
\begin{figure}[H]
    \captionsetup{justification=raggedright,singlelinecheck=false}

    \begin{minipage}[t]{0.485\linewidth}
        \vspace{0pt}
        \raggedleft
        \includegraphics[width=1\linewidth]{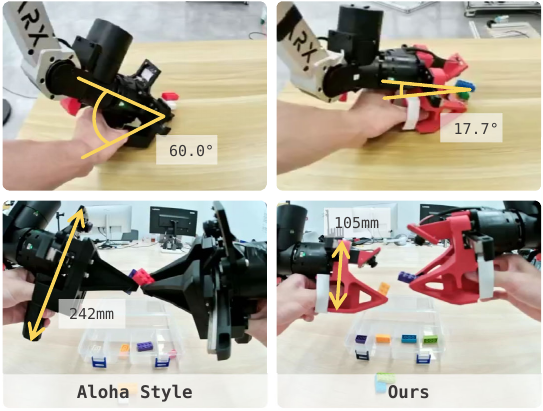}
        \caption{\textbf{Kinesthetic end-effector redesign.} Compared with the ALOHA end-effector, we remove the handle-and-scissor design for kinesthetic teaching. The handle limits access near the table surface, while the scissor mechanism is less convenient for forceful manipulation. Our design increases hand contact area, enabling compact and stable physical control.}
        \label{fig:EEF}
    \end{minipage}\hfill%
    \begin{minipage}[t]{0.485\linewidth}
        \vspace{-4pt}
        \raggedleft
        \includegraphics[width=0.9\linewidth]{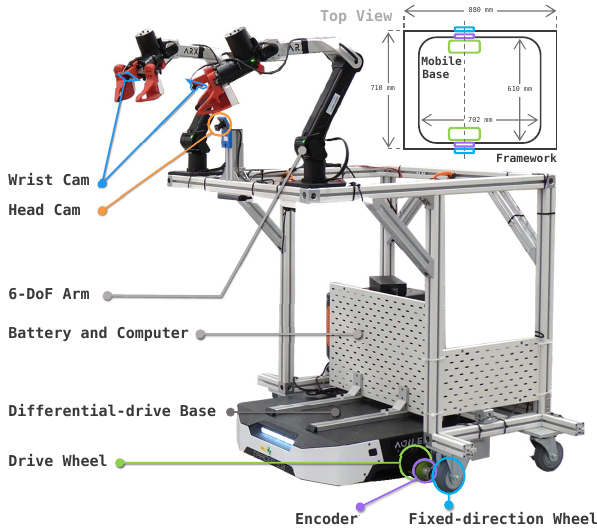}
        \vspace{-5pt}
        \caption{\textbf{\method{} deployment system overview.} Hardware configuration includes the dual-arm manipulation interface, onboard cameras, differential-drive base, encoders, and computing system. The base is removed during data collection.}
        \label{fig:car}
    \end{minipage}
\end{figure}

EgoBridge~\cite{punamiya2025egobridge} aligns human and robot demonstrations using dynamic time warping. Ego2Robot~\cite{wang2026ego2robot} first estimates the human hand pose and then replaces the hand with the robot arm through inpainting. Human2Sim2Robot~\cite{lum2025crossing} uses a single human demonstration together with reinforcement learning to derive robot-executable actions. In contrast, since our kinesthetic teaching interface directly provides ground-truth robot actions, we only need to mitigate demonstrator-induced artifacts in visual observations and control, without additional action annotation or replay.


\subsection{Learning-Based Mobile Manipulation}

Hierarchical models separate task-level manipulation prediction from whole-body execution, using planners or whole-body controllers to coordinate the base and arm~\cite{zhu2026emma,umi_on_legs,yuan2026hermes,xu2026hommi,zhao2025combined}. However, many real-world tasks require tight and continuous coordination between mobility and manipulation, making such modular designs less suitable for unified whole-body control and introducing additional system complexity. Unified policies~\cite{wu2024tidybot_1,pi05,mobile_aloha}, in contrast, directly model coupled base and manipulation actions, which is particularly effective when mobility and manipulation must evolve simultaneously. We find that simply concatenating base actions with arm actions already yields strong whole-body control without additional architectural design.

\section{Method}

HALO combines kinesthetic teaching with a low-resistance mobility interface to collect coordinated whole-body demonstrations. We first describe the manipulation interface in~\cref{sec:manipulation-interface} and the mobility interface in~\cref{sec:mobility-interface}. We then introduce deployment-aligned demonstration processing in~\cref{sec:deployment-aligned-processing}, including visual processing and base-motion alignment, to address the discrepancies caused by close human--robot coupling and base-free data collection. Finally, we describe policy learning and asynchronous execution in~\cref{sec:policy-learning}.

\halotopbreak
\halowraplabels{fig:demo,fig:mobile}
\begin{wrapfigure}[33]{R}{\halocolumnwidth}
    \raggedleft
    \captionsetup{justification=raggedright,singlelinecheck=false}
    \includegraphics[width=1\linewidth]{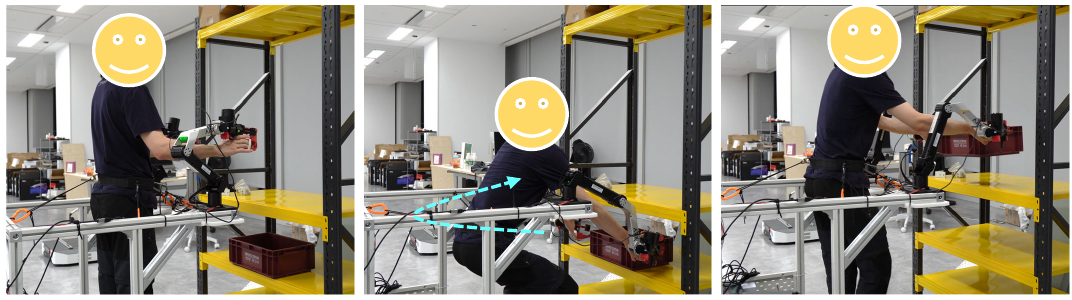}
    \caption{\textbf{Implicit base control.} The low-resistance collection platform allows base motion to arise naturally from the operator's movement, without explicit base control that could disrupt hand manipulation, thereby capturing coordinated mobile manipulation behaviors.}
    \label{fig:demo}

\medskip

    \raggedleft
    \captionsetup{justification=raggedright,singlelinecheck=false}
    \includegraphics[width=1.\linewidth]{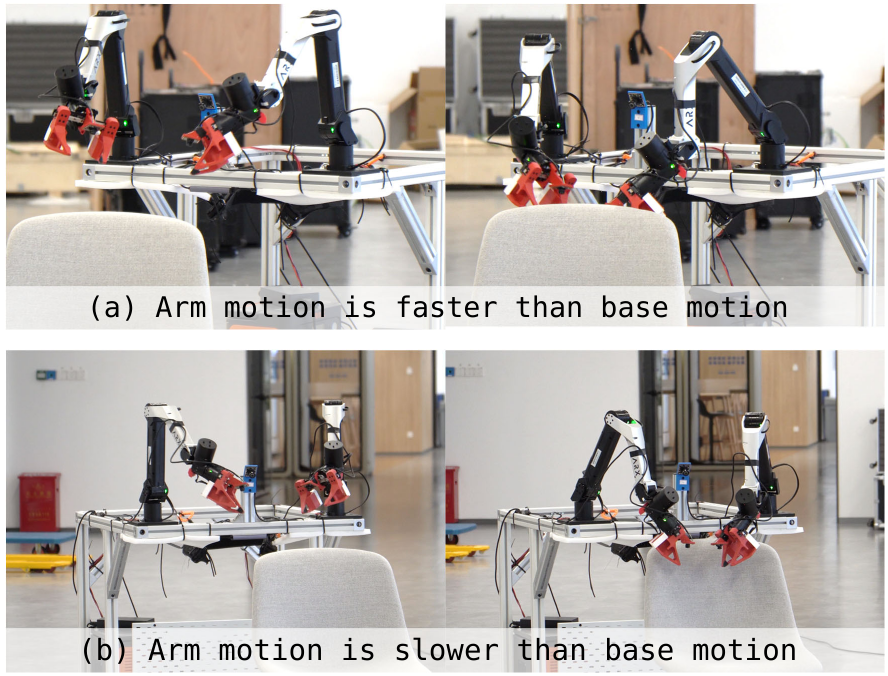}
    \caption{
\textbf{Failure cases in tasks requiring high mobility.}
(a) Without compensation for mobile-base latency, the arm motion advances faster than the base motion.
(b) Synchronous inference introduces additional control latency, causing the base to travel farther than intended.
}
    \label{fig:mobile}
\end{wrapfigure}

\halowrapheadingstart{0.10in}
\subsection{Manipulation Interface}\label{sec:manipulation-interface}

A data collection interface should adapt to the operator rather than require the operator to adapt to the robot. ALOHA's leader--follower design enables intuitive bimanual control, but the operator's hands remain spatially separated from the follower end-effectors, and the operator's egocentric view differs from the robot viewpoint. These gaps weaken perception--action alignment and direct robot control. We therefore use kinesthetic teaching to more closely align the operator's perception and action space with the robot.

As shown in~\cref{fig:EEF}, the original ALOHA end-effector is not designed for this type of physical interaction. Its handle limits the working space, while the scissor mechanism is less suitable for forceful manipulation. These features make stable grasping and force application less convenient in compact or contact-rich tasks.

So we redesign the end-effector for kinesthetic teaching by removing the handle-and-scissor mechanism and adopting a compact, high-contact design. As shown in~\cref{fig:EEF}, the redesign reduces the end-effector size by more than half and increases the available pitch angle by about $40^\circ$, enabling closer and more flexible manipulation while preserving direct visual and force feedback.

This close human--robot coupling makes bimanual demonstrations more natural, but it also causes the operator's hands to appear in the robot observations. We address this visual discrepancy in the deployment-aligned demonstration processing stage described in~\cref{sec:deployment-aligned-processing}.

\subsection{Mobility Interface}\label{sec:mobility-interface}

The mobility interface preserves natural operator locomotion while maintaining the motion constraints of the deployed base. Mobile ALOHA's powered base occupies the operator's movement space~\cite{mobile_aloha}. We remove the base during data collection and let natural body motion directly drive a low-resistance demonstration platform, as shown on the left of~\cref{fig:teaser} and~\cref{fig:demo}. This allows the operator to control mobility implicitly while focusing on bimanual manipulation.

The deployment system uses an AgileX Tracer AGV 2 differential-drive base, whereas the collection platform uses passive wheels and two fixed-direction wheels placed along the drive-wheel axis. This configuration mechanically constrains the collection platform to the same nonholonomic motion space as the deployed base, preserving kinematic compatibility without post-hoc retargeting.

\halotopbreak
\halowraplabels{fig:pipeline}
\begin{wrapfigure}[16]{R}{\halocolumnwidth}
    \raggedleft
    \captionsetup{justification=raggedright,singlelinecheck=false}
    \includegraphics[width=1\linewidth]{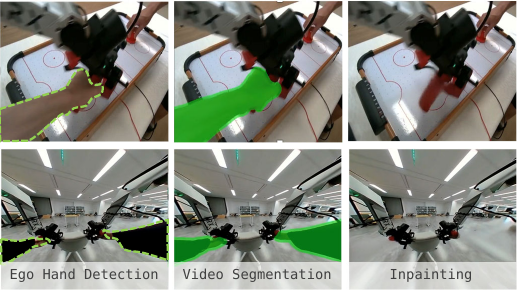}
    \caption{\textbf{Visual processing.} Our pipeline removes human hands during rapid motion, bimanual manipulation, and egocentric interaction.}
    \label{fig:pipeline}

\end{wrapfigure}

Removing the powered base also substantially reduces the force required for repositioning: moving the deployment system requires 47.3~N of horizontal force and Mobile ALOHA requires 13.0~N~\cite{mobile_aloha}, whereas our collection platform requires only 2.3~N. This reduction allows the operator to walk and reposition naturally, as shown in~\cref{fig:demo}.


Two encoders mounted near the drive-wheel axis measure the motion of the collection platform, as shown in the upper right of~\cref{fig:car}. Since the collection platform is passive and has different dynamics from the deployed base, the measured motion cannot be used directly as the deployment action. We therefore calibrate the mapping from platform motion to base velocity and estimate the response delay of the deployed base in~\cref{sec:base-motion-alignment}.

\subsection{Deployment-Aligned Demonstration Processing}\label{sec:deployment-aligned-processing}

The close human--robot interaction and base-free collection platform make demonstrations more natural, but they also introduce discrepancies between the demonstration and deployment settings. We address these discrepancies through deployment-aligned demonstration processing, covering both visual observations and base actions.

\subsubsection{Visual Processing}\label{sec:visual-processing}


The operator's hands are visible during kinesthetic teaching but absent during autonomous execution. To reduce this visual discrepancy, we process demonstration videos through hand segmentation, temporal mask propagation, and video inpainting, as shown in~\cref{fig:pipeline}.  We divide each video into overlapping temporal segments and sample frames within each segment to serve as keyframes. CaRE-Ego~\cite{su2025care} predicts left- and right-hand masks on these keyframes, which serve as prompts for SAM2~\cite{ravi2025sam}. Within each segment, we first propagate masks backward from the last keyframe with a valid hand detection, then forward from the first valid keyframe, covering frames before and after the detections. We blend mask probabilities where adjacent segments overlap to smooth boundary transitions, then remove small isolated components, fill holes, and dilate the masks to cover residual hand boundaries. DiffuEraser~\cite{li2025diffueraser} then combines optical-flow-guided propagation of visible scene content with diffusion-based inpainting to reconstruct the masked regions. Finally, we blend the repaired regions with the original frames. The resulting observations suppress demonstrator-specific cues while retaining the surrounding scene, reducing the visual discrepancy between demonstration and deployment.

\subsubsection{Base Motion Alignment}\label{sec:base-motion-alignment}
With the kinematic constraints already matched by the collection interface, the remaining alignment accounts for differences in dynamics and control response between the passive platform and the powered base.
Encoder readings are converted into left and right platform wheel velocities, $u_L$ and $u_R$. We characterize the response of the deployed base from commanded velocities to encoder-observed motion and estimate the resulting action--state delay for temporal alignment, following~\cite{mobile_aloha}. Ignoring this delay introduces a systematic mismatch between commanded actions and observed base motion, as illustrated in~\cref{fig:mobile}. We further fit an affine mapping from the measured platform motion to deployment-compatible base velocities:

\begin{equation}
\begin{bmatrix}
v \\
\omega
\end{bmatrix}
=
\begin{bmatrix}
k_{vL} & k_{vR} \\
k_{\omega L} & k_{\omega R}
\end{bmatrix}
\begin{bmatrix}
u_L \\
u_R
\end{bmatrix}
+
\begin{bmatrix}
d_v \\
d_\omega
\end{bmatrix},
\end{equation}

where $v$ and $\omega$ are the deployment-compatible linear and angular velocities. The coefficients and offsets are estimated from paired platform measurements and commanded base velocities collected on the deployment system. To avoid compounding errors from integrating estimated motion, we use direct base velocity rather than pose as the action representation.

\subsection{Policy Learning}\label{sec:policy-learning}

To evaluate the effectiveness of the demonstrations collected with \method{}, we use the pretrained $\pi_{0.5}$ vision-language-action model~\cite{pi05} as the base policy and fine-tune it on our demonstrations. The policy receives three images, including one head-camera image and two wrist-camera images, together with the robot joint positions and base linear and angular velocities. The action vector concatenates the arm actions with the base linear and angular velocities. For each task, we fine-tune the pretrained policy with AdamW using a constant learning rate of $5\times10^{-5}$, a batch size of 128, and 5k training steps. 
Models trained without pretrained initialization use the same training setup for 10k steps.

Tight whole-body coordination requires continuous policy execution with frequent replanning. We therefore use asynchronous real-time action chunking~(RTC)~\cite{rtc}, where the policy predicts the next action chunk while the current chunk is being executed. Given the latest observation $o_t$ and the committed action prefix $a_{t+d-1}$, the policy predicts the next action chunk:
\begin{equation}
a_{t+d:t+d+H-1}\sim\pi(\cdot\mid o_t,a_{t+d-1}),
\end{equation}
where $d$ is the inference delay in control steps and $H$ is the prediction horizon. Conditioning on the committed action prefix maintains continuity between consecutive action chunks. We further ablate the effects of inference settings, pretraining, and action representation on policy performance.


\section{Evaluations}

We evaluate our framework along three key dimensions that reflect its ability to preserve human manipulation skills while enabling efficient and coordinated whole-body policy learning:

\begin{itemize}
    \item \textbf{Data collection efficiency.} How efficiently can users collect successful whole-body demonstrations with our system compared with existing teleoperation interfaces?
    
    \item \textbf{Dexterity.} How effectively can demonstrations collected with our system transfer human manipulation dexterity to learned robot policies, particularly for precise and bimanual manipulation?
    
    \item \textbf{Coordinated whole-body control.} How well can the learned policy coordinate base motion and manipulation, especially for tasks requiring precise repositioning and long-horizon execution?
\end{itemize}

For all experimental tasks, we limit the complete data collection process to one hour, including scene reset and object preparation. We impose no constraints on demonstration variability, such as fixed initial base poses or object placements. Operators are instructed to collect demonstrations at a natural pace and to walk naturally during mobile manipulation, allowing us to evaluate whether \method{} can capture natural human whole-body motion.

\subsection{Data Collection Efficiency}

We compare \method{} with VR-based teleoperation and Mobile ALOHA in terms of demonstration success and data collection efficiency, focusing on tasks that require tight coordination between mobility and bimanual manipulation. For a fair comparison, all methods are evaluated by the same operator under the same task settings, collection time budget, and reset protocol. The operator had no prior experience with any of the evaluated interfaces before the study.

\halowrapfinish
\halowraplabels{fig:VR_failure}
\begin{wrapfigure}{R}{\halocolumnwidth}

    \raggedleft
    \captionsetup{justification=raggedright,singlelinecheck=false}
    \includegraphics[width=1\linewidth]{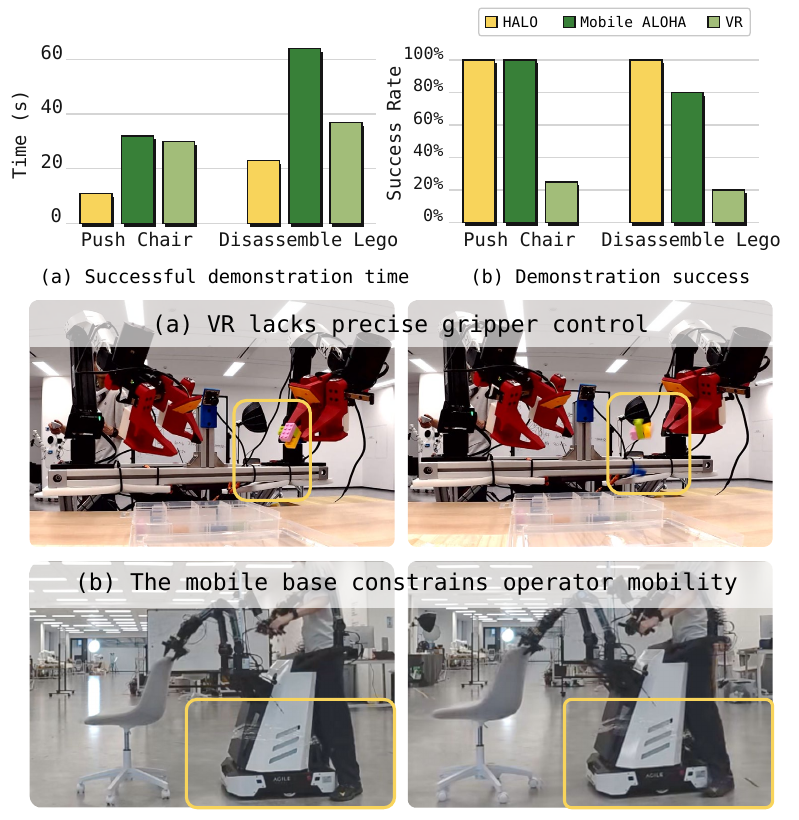}
    \caption{\textbf{Comparison and limitations of demonstration data collection.}
\textbf{Top:} Comparison of human demonstration collection methods in terms of efficiency and reliability.
Succ. Rate and Time denote the demonstration success rate and the average duration of successful demonstrations, respectively.
\textbf{Bottom:} Representative limitations of existing teleoperation interfaces: VR lacks precise gripper control and force feedback, while the mobile base constrains operator mobility during whole-body demonstration collection.}
    \label{fig:VR_failure}
\end{wrapfigure}

\noindent \textbf{Task description.}
We evaluate two tasks, \textit{Push Chair} and \textit{Disassemble Lego}. In \textit{Push Chair} (\cref{fig:zip_tie_lego}), the operator grasps the chair back with both grippers while moving at approximately normal walking speed ($\sim 0.7\,\mathrm{m/s}$), then navigates to push the chair under a table. This task requires stable bimanual grasping during rapid base motion. In \textit{Disassemble Lego} (\cref{fig:zip_tie_lego}), the operator disassembles a four-piece Lego structure and places each piece into the bin of the corresponding color. The Lego configuration and object arrangement are randomized, requiring precise and adaptive bimanual manipulation.

\noindent \textbf{Results.}
\method{} matches or exceeds the baselines in demonstration success rate while reducing the average duration of successful demonstrations by up to 66\%. Compared with Mobile ALOHA, \method{} achieves speedups of $2.9\times$ on \textit{Push Chair} and $2.8\times$ on \textit{Disassemble Lego}, while improving the latter's demonstration success rate by 20 percentage points.
As illustrated in~\cref{fig:VR_failure}, the two baselines present distinct interaction difficulties: VR's perception gap hinders precise gripper control, whereas Mobile ALOHA's physical base constrains operator movement. On \textit{Push Chair}, this constraint leads the operator to pause before grasping, increasing demonstration time despite a high success rate. \method{} addresses these difficulties through direct manipulation control and a low-resistance platform that allows the operator to reposition naturally.

\WFclear
\subsection{Dexterity}


\halowraplabels{tab:Dexterity}
\begin{wraptable}{R}{\halocolumnwidth}
\vspace*{-9.8pt}
\centering
\captionsetup{justification=raggedright,singlelinecheck=false}
\caption{\textbf{Dexterity evaluation.} Zip-Tie Fastening is evaluated over 10 trials. Succ. Rate denotes the task success rate, and Time denotes the average completion time.}
\label{tab:Dexterity}
\begin{adjustbox}{max width=1\linewidth}
\begin{tabular}{cccccc}
\toprule
\multicolumn{6}{c}{Zip-Tie Fastening} \\
\cmidrule(lr){1-6}
\multicolumn{2}{c}{\method}
& \multicolumn{2}{c}{ALOHA}
& \multicolumn{2}{c}{VR} \\
\cmidrule(lr){1-2}
\cmidrule(lr){3-4}
\cmidrule(lr){5-6}
Succ. Rate & Time (s)
& Succ. Rate & Time (s)
& Succ. Rate & Time (s) \\
\midrule
30.0\% & 41
& 0.0\% & -
& 0.0\% & - \\
\bottomrule
\end{tabular}
\end{adjustbox}
\end{wraptable}
We evaluate whether \method{} can capture tasks requiring precise manipulator and base control, and whether the learned policy can reproduce high-precision behaviors.

\begin{figure}[H]

    \centering
    \includegraphics[width=1\linewidth]{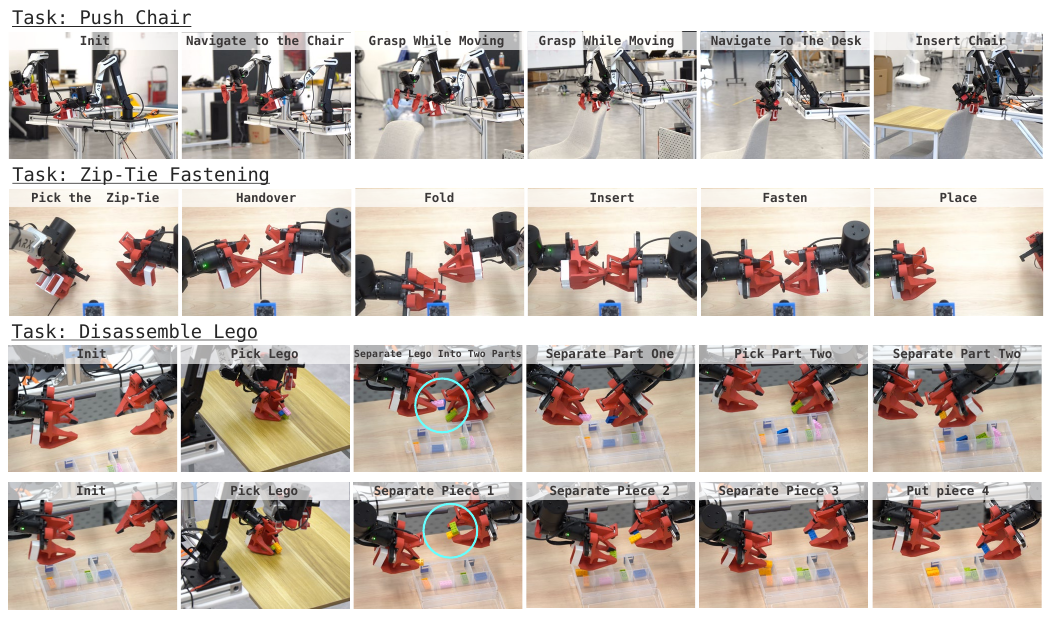}
    \caption{\textbf{Push Chair, Zip-Tie Fastening, and Disassemble Lego.}
    In Push Chair, the policy navigates to the chair, grasps it while moving, transports it to the target desk, and inserts it into place. 
    Both Zip-Tie Fastening and Disassemble Lego require precise action prediction. In Disassemble Lego, the randomized separation pattern changes the required disassembly strategy. The policy learns to handle these variations from our demonstrations and generalizes to unseen Lego configurations. In the third row, two Lego pieces of the same color are included, and the policy correctly places both resulting pieces into the correct bin.}
    \label{fig:zip_tie_lego}

\end{figure}

\noindent \textbf{Task description.} As shown in~\cref{fig:zip_tie_lego}, \textit{Zip-Tie Fastening} requires a sequence of precise bimanual operations, including picking, handover, folding, insertion, fastening, and placement. The policy must manipulate a short, small zip tie throughout this process, requiring sub-millimeter manipulation accuracy.


\noindent \textbf{Results.} As shown in~\cref{fig:zip_tie_lego} and~\cref{tab:Dexterity}, the policy successfully reproduces the precise manipulation required for Zip-Tie Fastening. In contrast, Mobile ALOHA suffers from the perception gap, as the operator remains too far from the zip tie to clearly observe the fine manipulation details. VR teleoperation cannot reliably achieve the control precision and dexterity required for folding and fastening the zip tie. As a result, both interfaces yield low demonstration success rates and poor-quality data, preventing the learned policies from successfully completing the task. 



\halowrapfinish

\subsection{Coordinated whole-body control}
We test \method{} on tasks that require both precise whole-body coordination and long-horizon mobile manipulation.

\noindent \textbf{Task description.}  We evaluate two tasks: In \textit{Disassemble Lego}, the policy must precisely position the base relative to the workspace and coordinate both end-effectors to grasp, separate, and place small interlocked Lego pieces. Varying connection strengths require coordinated bimanual force and motion control together with accurate base positioning, as shown in~\cref{fig:zip_tie_lego}. 

\begin{figure}[H]

    \centering
    \includegraphics[width=1\linewidth]{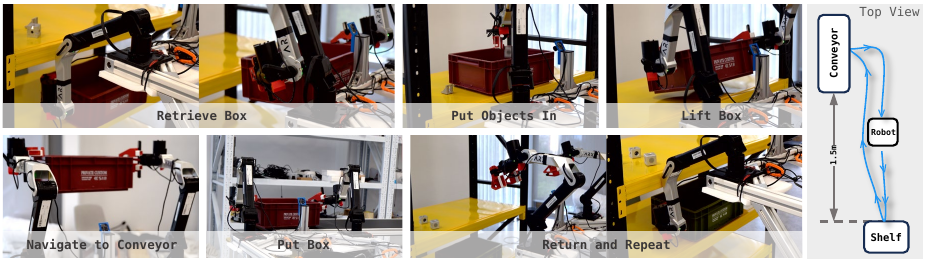}
    \caption{\textbf{Long-horizon mobile manipulation with coordinated whole-body control.} The policy retrieves a box from the shelf, loads target objects, transports the box to the conveyor, places it at the designated location, and returns to repeat the procedure.}
    \label{fig:box_2_conveyor}

\end{figure}

For the long-horizon task, the procedure consists of six stages as shown in~\cref{fig:box_2_conveyor}. The policy first retrieves a box from the shelf, then places target objects into the box. It next lifts the loaded box with both arms, navigates to the conveyor, and places the box at the designated location. Finally, the robot returns to the shelf and repeats the procedure. The task requires coordinated bimanual manipulation, accurate base navigation, and reliable long-horizon execution across an extended workspace.

\noindent \textbf{Results.}
In \textit{Disassemble Lego}, the policy succeeds in 30 out of 40 trials (75.0\%). For the long-horizon box manipulation task, the policy succeeds in 11 out of 15 trials (73.3\%). Despite multiple forward, backward, and rotational base motions throughout the task, the policy consistently returns the mobile base to its initial position, enabling the full procedure to be repeated reliably. For both VR and Mobile ALOHA, policies trained with the same data collection time budget fail on both tasks, due to slow demonstrations with frequent pauses and insufficient valid data resulting from high collection failure rates.


\begin{table}[H]
\captionsetup{justification=raggedright,singlelinecheck=false}
\begin{minipage}[t]{0.58\linewidth}
\vspace{0pt}
\centering
\caption{\textbf{Inference Strategy.} Success rates on the \textit{Push Chair} task under synchronous and asynchronous inference with different execution horizons.}
\vspace{10pt}
\begingroup
\setlength{\tabcolsep}{3pt}
\begin{adjustbox}{max width=\linewidth}
    \begin{tabular}{lcccccc}
    \toprule
    Push Chair        & \multicolumn{2}{c}{Synchronous} & \multicolumn{4}{c}{Asynchronous} \\
    \cmidrule(lr){1-1}
    \cmidrule(lr){2-3}
    \cmidrule(lr){4-7}
    Execution Horizon  & 8       & 25     & 4      & 6      & 8      & 10     \\ \midrule
    Success Rate       & 0.0\%  & 35.0\%  & 75.0\% & 75.0\% & 85.0\% & 70.0\% \\
    \bottomrule
    \end{tabular}
\end{adjustbox}
\endgroup
\label{tab:inference}
\end{minipage}\hfill%
\begin{minipage}[t]{0.39\linewidth}
\vspace{0pt}
\centering
\caption{\textbf{Action representation ablation.} Success rates on the \textit{Disassemble Lego} task for different action spaces and base models.}
\begin{adjustbox}{max width=\linewidth}
\begin{tabular}{lcc}
\toprule
& From VLM & Pretrained VLA \\
\midrule
EEF   & 12.5\% & 47.5\% \\
Joint & 7.5\% & 75.0\% \\
\bottomrule
\end{tabular}
\end{adjustbox}
\label{tab:action}
\end{minipage}
\end{table}

\subsection{Ablation Studies}


\noindent \textbf{Inference strategies.} Since tight whole-body coordination is sensitive to inference pauses, we further study the effect of RTC-specific execution design on policy performance, as shown in~\Cref{tab:inference}. Under \textit{synchronous execution}, inference pauses cannot be fully eliminated, and longer execution horizons yield higher success rates because frequent replanning introduces more pauses and disrupts coordination between the arms and mobile base.  \textit{Asynchronous execution} achieves higher overall success by mitigating the effect of inference latency. However, more frequent replanning does not always improve performance, indicating a trade-off between reactivity and temporal consistency. Frequent replanning can reduce temporal consistency across action chunks, while longer horizons reduce responsiveness to new observations. Our results show that an intermediate replanning horizon provides the best balance.



\halowraplabels{fig:chair2desk}
\begin{wrapfigure}[20]{R}{\halocolumnwidth}
    \raggedleft
    \captionsetup{justification=raggedright,singlelinecheck=false}
    \includegraphics[width=1.\linewidth]{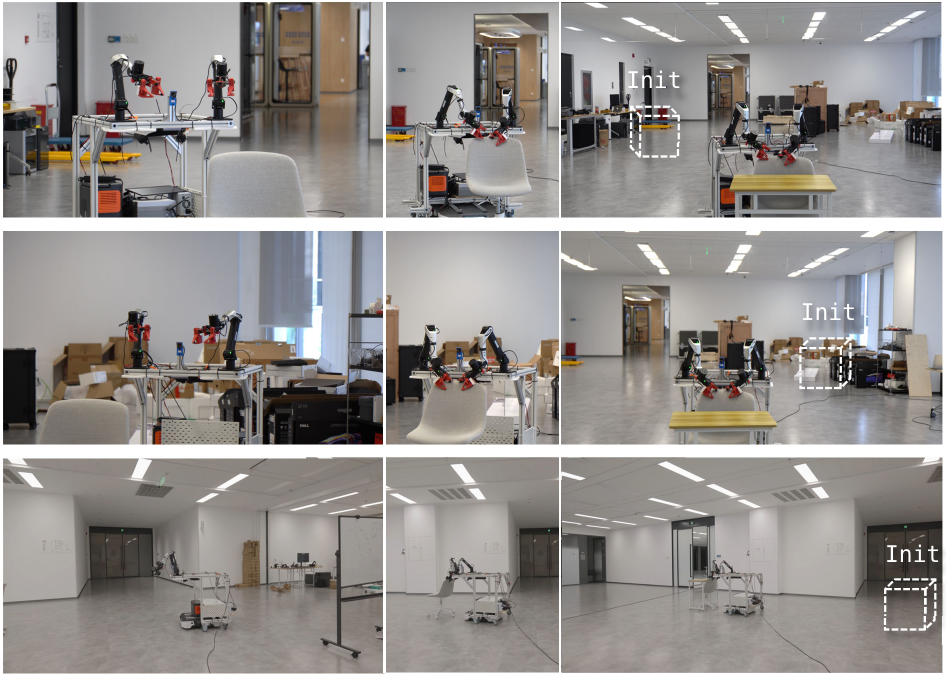}
    \caption{\textbf{Generalization of mobile manipulation.} We evaluate \textit{Push Chair} under large initial-pose variations and in unseen environments. Compared with the 5--10\,cm initial-position variation in Mobile ALOHA, we increase the randomization to 2--3\,m.}
    \label{fig:chair2desk}
\end{wrapfigure}

\noindent \textbf{Deployment-aligned visual post-processing.}
We evaluate whether removing demonstrator-specific visual cues improves alignment with autonomous deployment. In \textit{Push Chair}, we compare policies trained on the same demonstrations with and without visual processing in~\cref{sec:visual-processing}, using the best inference setting~(asynchronous inference with an execution horizon length of 8). Hand removal improves the success rate from 55\% to 85\%, a gain of 30 percentage points, demonstrating the effectiveness of deployment-aligned visual processing. 

\noindent \textbf{Action representation.}
We compare end-effector (EEF) and joint-space actions for policy training on the \textit{Disassemble
Lego} task. We convert recorded joint-space actions into EEF actions using forward kinematics with the official ARX-X5 robot URDF, and convert predicted EEF actions back into joint-space commands using Mink~\cite{Zakka_Mink_Python_inverse_2026} for execution.
As shown in~\cref{tab:action}, EEF performs better when trained from scratch ($\pi_{0.5}$ architecture with PaliGemma~\cite{beyer2024paligemma} initialization), while joint-space actions achieve higher success rates when using the pretrained $\pi_{0.5}$ checkpoint.
This suggests that pretraining enables the policy to better exploit direct joint-space control.

\newpage

\noindent \textbf{Generalization of mobile manipulation.}
Although the target table remains fixed during data collection, the learned VLA policy generalizes \textit{Push Chair }to unseen backgrounds and substantially different chair--table configurations, as shown in~\cref{fig:chair2desk}. We attribute this generalization partly to velocity-based base control, which encourages the policy to learn a reactive motion pattern that continuously drives the chair toward the table rather than memorizing a fixed global pose.


\halowrapheadingstart{0.12in}




\section{Conclusion}

In this work, we presented \method{}, a human-aligned data collection system for whole-body mobile manipulation. Our key insight is that the robot interface should adapt to human capabilities by aligning perception and manipulation while preserving natural locomotion. This design enables efficient collection of coordinated demonstrations, while demonstrator decoupling mitigates the visual and control shifts introduced by close human--robot coupling. Building on this alignment, future work could incorporate active perception~\cite{xiong2025vision,xu2026hommi} to adapt the robot viewpoint during manipulation and more closely reflect human visual behavior. Across rapid mobile manipulation, dexterous tasks, and long-horizon behaviors, \method{} demonstrates that human-aligned interfaces provide an effective foundation for efficient data collection and robust whole-body policy learning.

\halowrapfinish
\bibliography{references}

@article{umi,
  title={Universal manipulation interface: In-the-wild robot teaching without in-the-wild robots},
  author={Chi, Cheng and Xu, Zhenjia and Pan, Chuer and Cousineau, Eric and Burchfiel, Benjamin and Feng, Siyuan and Tedrake, Russ and Song, Shuran},
  journal={arXiv preprint arXiv:2402.10329},
  year={2024}
}

@article{aloha,
  title={Learning fine-grained bimanual manipulation with low-cost hardware},
  author={Zhao, Tony Z and Kumar, Vikash and Levine, Sergey and Finn, Chelsea},
  journal={arXiv preprint arXiv:2304.13705},
  year={2023}
}

@article{mobile_aloha,
  title={Mobile aloha: Learning bimanual mobile manipulation with low-cost whole-body teleoperation},
  author={Fu, Zipeng and Zhao, Tony Z and Finn, Chelsea},
  journal={arXiv preprint arXiv:2401.02117},
  year={2024}
}

@article{xu2026hommi,
  title={Hommi: Learning whole-body mobile manipulation from human demonstrations},
  author={Xu, Xiaomeng and Park, Jisang and Zhang, Han and Cousineau, Eric and Bhat, Aditya and Barreiros, Jose and Wang, Dian and Bohg, Jeannette and Song, Shuran},
  journal={arXiv preprint arXiv:2603.03243},
  year={2026}
}

@inproceedings{akgun2012trajectories,
  title={Trajectories and keyframes for kinesthetic teaching: A human-robot interaction perspective},
  author={Akgun, Baris and Cakmak, Maya and Yoo, Jae Wook and Thomaz, Andrea Lockerd},
  booktitle={Proceedings of the seventh annual ACM/IEEE international conference on Human-Robot Interaction},
  pages={391--398},
  year={2012}
}

@inproceedings{gello,
  title={Gello: A general, low-cost, and intuitive teleoperation framework for robot manipulators},
  author={Wu, Philipp and Shentu, Yide and Yi, Zhongke and Lin, Xingyu and Abbeel, Pieter},
  booktitle={2024 IEEE/RSJ International Conference on Intelligent Robots and Systems (IROS)},
  pages={12156--12163},
  year={2024},
  organization={IEEE}
}

@article{zhu2022bottom,
  title={Bottom-up skill discovery from unsegmented demonstrations for long-horizon robot manipulation},
  author={Zhu, Yifeng and Stone, Peter and Zhu, Yuke},
  journal={IEEE Robotics and Automation Letters},
  volume={7},
  number={2},
  pages={4126--4133},
  year={2022},
  publisher={IEEE}
}

@inproceedings{rakita2018autonomous,
  title={An autonomous dynamic camera method for effective remote teleoperation},
  author={Rakita, Daniel and Mutlu, Bilge and Gleicher, Michael},
  booktitle={Proceedings of the 2018 ACM/IEEE International Conference on Human-Robot Interaction},
  pages={325--333},
  year={2018}
}

@article{yu2025egomi,
  title={Egomi: Learning active vision and whole-body manipulation from egocentric human demonstrations},
  author={Yu, Justin and Shentu, Yide and Wu, Di and Abbeel, Pieter and Goldberg, Ken and Wu, Philipp},
  journal={arXiv preprint arXiv:2511.00153},
  year={2025}
}

@article{citron2026modpack,
  title={ModPack: An Extensible Teleoperation Interface for Bimanual Mobile Manipulation},
  author={Citron, Joshua and Zbizika, Renee and Liu, Zeyi and Song, Shuran},
  journal={arXiv preprint arXiv:2607.19479},
  year={2026}
}

@article{honerkamp2025whole,
  title={Whole-body teleoperation for mobile manipulation at zero added cost},
  author={Honerkamp, Daniel and Mahesheka, Harsh and von Hartz, Jan Ole and Welschehold, Tim and Valada, Abhinav},
  journal={IEEE Robotics and Automation Letters},
  volume={10},
  number={4},
  pages={3198--3205},
  year={2025},
  publisher={IEEE}
}

@article{liang2026universal,
  title={Universal Manipulation Exoskeleton: Learning Compliant Whole-body Policies with Real-time Torque Feedback},
  author={Liang, Litian and Xu, Jingxi and Qi, Xinda and Cai, Yujun and Ding, Houzhu and Wang, Luqi and Sun, Zhixin and Chow, Jyh-Herng and Yang, Ming and Cutkosky, Mark},
  journal={arXiv preprint arXiv:2606.14218},
  year={2026}
}

@article{black2024pi_0,
  title={$\pi_0$: A Vision-Language-Action Flow Model for General Robot Control},
  author={Black, Kevin and Brown, Noah and Driess, Danny and Esmail, Adnan and Equi, Michael and Finn, Chelsea and Fusai, Niccolo and Groom, Lachy and Hausman, Karol and Ichter, Brian and others},
  journal={arXiv preprint arXiv:2410.24164},
  year={2024}
}

@article{dp,
  title={Diffusion policy: Visuomotor policy learning via action diffusion},
  author={Chi, Cheng and Xu, Zhenjia and Feng, Siyuan and Cousineau, Eric and Du, Yilun and Burchfiel, Benjamin and Tedrake, Russ and Song, Shuran},
  journal={The International Journal of Robotics Research},
  volume={44},
  number={10-11},
  pages={1684--1704},
  year={2025},
  publisher={Sage Publications Sage UK: London, England}
}

@article{kim2024openvla,
  title={Openvla: An open-source vision-language-action model},
  author={Kim, Moo Jin and Pertsch, Karl and Karamcheti, Siddharth and Xiao, Ted and Balakrishna, Ashwin and Nair, Suraj and Rafailov, Rafael and Foster, Ethan and Lam, Grace and Sanketi, Pannag and others},
  journal={arXiv preprint arXiv:2406.09246},
  year={2024}
}

@article{chen2026supersuit,
  title={SuperSuit: An Isomorphic Bimodal Interface for Scalable Mobile Manipulation},
  author={Chen, Tongqing and Wu, Hang and Wang, Jiasen and Li, Xiaotao and Jin, Zhu and Fang, Lu},
  journal={arXiv preprint arXiv:2603.06280},
  year={2026}
}

@article{jiang2025behavior,
  title={Behavior robot suite: Streamlining real-world whole-body manipulation for everyday household activities},
  author={Jiang, Yunfan and Zhang, Ruohan and Wong, Josiah and Wang, Chen and Ze, Yanjie and Yin, Hang and Gokmen, Cem and Song, Shuran and Wu, Jiajun and Fei-Fei, Li},
  journal={arXiv preprint arXiv:2503.05652},
  year={2025}
}

@article{zhu2026emma,
  title={Emma: Scaling mobile manipulation via egocentric human data},
  author={Zhu, Lawrence Y and Kuppili, Pranav and Punamiya, Ryan and Aphiwetsa, Patcharapong and Patel, Dhruv and Kareer, Simar and Ha, Sehoon and Xu, Danfei},
  journal={IEEE Robotics and Automation Letters},
  year={2026},
  publisher={IEEE}
}

@article{umi_on_legs,
  title={UMI on Legs: Making Manipulation Policies Mobile with Manipulation-Centric Whole-body Controllers},
  author={Ha, Huy and Gao, Yihuai and Fu, Zipeng and Tan, Jie and Song, Shuran},
  journal={arXiv preprint arXiv:2407.10353},
  year={2024},
  doi={10.48550/arXiv.2407.10353}
}

@article{yuan2026hermes,
  title={Hermes: Human-to-robot embodied learning from multi-source motion data for mobile dexterous manipulation},
  author={Yuan, Zhecheng and Wei, Tianming and Gu, Langzhe and Hua, Pu and Liang, Tianhai and Chen, Yuanpei and Xu, Huazhe},
  journal={IEEE Transactions on Robotics},
  year={2026},
  publisher={IEEE}
}

@article{pi05,
  title={$\pi_{0.5}$: a Vision-Language-Action Model with Open-World Generalization},
  author={Intelligence, Physical and Black, Kevin and Brown, Noah and Darpinian, James and Dhabalia, Karan and Driess, Danny and Esmail, Adnan and Equi, Michael and Finn, Chelsea and Fusai, Niccolo and others},
  journal={arXiv preprint arXiv:2504.16054},
  year={2025}
}

@article{su2025care,
  title={Care-ego: Contact-aware relationship modeling for egocentric interactive hand-object segmentation},
  author={Su, Yuejiao and Wang, Yi and Chau, Lap-Pui},
  journal={Expert Systems with Applications},
  pages={129148},
  year={2025},
  publisher={Elsevier}
}

@inproceedings{ravi2025sam,
  title={Sam 2: Segment anything in images and videos},
  author={Ravi, Nikhila and Gabeur, Valentin and Hu, Yuan-Ting and Hu, Ronghang and Ryali, Chaitanya and Ma, Tengyu and Khedr, Haitham and R{\"a}dle, Roman and Rolland, Chloe and Gustafson, Laura and others},
  booktitle={International Conference on Learning Representations},
  volume={2025},
  pages={28085--28128},
  year={2025}
}

@article{li2025diffueraser,
  title={Diffueraser: A diffusion model for video inpainting},
  author={Li, Xiaowen and Xue, Haolan and Ren, Peiran and Bo, Liefeng},
  journal={arXiv preprint arXiv:2501.10018},
  year={2025}
}

@article{beyer2024paligemma,
  title={Paligemma: A versatile 3b vlm for transfer},
  author={Beyer, Lucas and Steiner, Andreas and Pinto, Andr{\'e} Susano and Kolesnikov, Alexander and Wang, Xiao and Salz, Daniel and Neumann, Maxim and Alabdulmohsin, Ibrahim and Tschannen, Michael and Bugliarello, Emanuele and others},
  journal={arXiv preprint arXiv:2407.07726},
  year={2024}
}

@article{rtc,
  title={Training-time action conditioning for efficient real-time chunking},
  author={Black, Kevin and Ren, Allen Z and Equi, Michael and Levine, Sergey},
  journal={arXiv preprint arXiv:2512.05964},
  year={2025}
}

@inproceedings{punamiya2025egobridge,
  title={Egobridge: Domain adaptation for generalizable imitation from egocentric human data},
  author={Punamiya, Ryan and Patel, Dhruv and Aphiwetsa, Patcharapong and Kuppili, Pranav and Zhu, Lawrence Y and Kareer, Simar and Hoffman, Judy and Xu, Danfei},
  booktitle={Human to Robot: Workshop on Sensorizing, Modeling, and Learning from Humans},
  year={2025}
}

@article{wang2026ego2robot,
  title={Ego2Robot: Scalable Robot Data Synthesis from Egocentric Human Data},
  author={Wang, Ye and Lin, Pei and Chen, Xiong-Hui and Yuan, Haoqi and Liang, Zhixuan and Huang, Yiyang and Chen, Anzhe and Lei, Zixing and Zhang, Jie and Zhang, Tao and others},
  journal={arXiv preprint arXiv:2608.02580},
  year={2026}
}

@article{xu2025dexumi,
  title={Dexumi: Using human hand as the universal manipulation interface for dexterous manipulation},
  author={Xu, Mengda and Zhang, Han and Hou, Yifan and Xu, Zhenjia and Fan, Linxi and Veloso, Manuela and Song, Shuran},
  journal={arXiv preprint arXiv:2505.21864},
  year={2025}
}

@article{va,
  title={Causal world modeling for robot control},
  author={Li, Lin and Zhang, Qihang and Luo, Yiming and Yang, Shuai and Wang, Ruilin and Han, Fei and Yu, Mingrui and Gao, Zelin and Xue, Nan and Zhu, Xing and others},
  journal={arXiv preprint arXiv:2601.21998},
  year={2026}
}

@article{vla,
  title={A pragmatic vla foundation model},
  author={Wu, Wei and Lu, Fan and Wang, Yunnan and Yang, Shuai and Liu, Shi and Wang, Fangjing and Zhu, Qian and Sun, He and Wang, Yong and Ma, Shuailei and others},
  journal={arXiv preprint arXiv:2601.18692},
  year={2026}
}

@article{xiong2025vision,
  title={Vision in action: Learning active perception from human demonstrations},
  author={Xiong, Haoyu and Xu, Xiaomeng and Wu, Jimmy and Hou, Yifan and Bohg, Jeannette and Song, Shuran},
  journal={arXiv preprint arXiv:2506.15666},
  year={2025}
}

@article{lum2025crossing,
  title={Crossing the human-robot embodiment gap with sim-to-real rl using one human demonstration},
  author={Lum, Tyler Ga Wei and Lee, Olivia Y and Liu, C Karen and Bohg, Jeannette},
  journal={arXiv preprint arXiv:2504.12609},
  year={2025}
}

@article{wu2024tidybot_1,
  title={Tidybot++: An open-source holonomic mobile manipulator for robot learning},
  author={Wu, Jimmy and Chong, William and Holmberg, Robert and Prasad, Aaditya and Gao, Yihuai and Khatib, Oussama and Song, Shuran and Rusinkiewicz, Szymon and Bohg, Jeannette},
  journal={arXiv preprint arXiv:2412.10447},
  year={2024}
}

@inproceedings{wong2022error,
  title={Error-aware imitation learning from teleoperation data for mobile manipulation},
  author={Wong, Josiah and Tung, Albert and Kurenkov, Andrey and Mandlekar, Ajay and Fei-Fei, Li and Savarese, Silvio and Mart{\'\i}n-Mart{\'\i}n, Roberto},
  booktitle={Conference on Robot Learning},
  pages={1367--1378},
  year={2022},
  organization={PMLR}
}

@article{zhao2025combined,
  title={A combined learning and optimization framework to transfer human whole-body loco-manipulation skills to mobile manipulators},
  author={Zhao, Jianzhuang and Tassi, Francesco and Huang, Yanlong and De Momi, Elena and Ajoudani, Arash},
  journal={Robotics and Autonomous Systems},
  volume={189},
  pages={104958},
  year={2025},
  publisher={Elsevier}
}

@software{Zakka_Mink_Python_inverse_2026,
  author = {Zakka, Kevin},
  title = {{Mink: Python inverse kinematics based on MuJoCo}},
  year = {2026},
  month = oct,
  version = {1.4.0},
  url = {https://github.com/kevinzakka/mink},
  license = {Apache-2.0}
}

@inproceedings{zhao2026xrobotoolkit,
  title     = {{XRoboToolkit}: A Cross-Platform Framework for Robot Teleoperation},
  author    = {Zhao, Zhigen and Yu, Liuchuan and Jing, Ke and Yang, Ning},
  booktitle = {2026 IEEE/SICE International Symposium on System Integration (SII)},
  year      = {2026},
  pages     = {15--20},
  doi       = {10.1109/SII64115.2026.11404528}
}

@article{shen2026fomi,
  title={FOMI: A Force-Feedback Manipulation Interface for Teaching Contact-Rich Manipulation},
  author={Shen, Yishu and Luo, Yiming and Wang, Ruilin and Yang, Bowen and Yang, Shuai and Xu, Fangyi and Shen, Yujun and Zhang, Fu and Zhou, Boyu and Xu, Yinghao},
  journal={arXiv preprint arXiv:2610.19839},
  year={2026}
}
\bibliographystyle{icml2025}

\clearpage
\appendix
\startcontents[appendix]


\begin{center}
    {\LARGE\bfseries Appendix\par}
\end{center}
\vspace{1em}

\section{Baseline Implementations}
\label{app:baseline-implementations}

\noindent\textbf{VR teleoperation.}
We use XRoboToolkit~\cite{zhao2026xrobotoolkit} to receive pose updates from the VR controllers. For each arm, we map the controller's relative translation and rotation to a relative motion expressed in the robot end-effector frame. We use the resulting target pose to solve inverse kinematics and send the joint commands to the arms.

\noindent\textbf{Mobile ALOHA.}
We use the AgileX Robotics setup for our Mobile ALOHA baseline~\cite{mobile_aloha}. During demonstration collection, we power off the mobile base so that the operator can move it manually while teleoperating the arms through the leader arms. External encoders measure the resulting base motion, which can be used to construct the base action labels.

\section{Training configuration.}

\begin{table}[h]
\centering
\caption{Policy training configuration.}
\label{tab:training-config}
\begin{tabular}{ll}
\toprule
Parameter & Value \\
\midrule
Base model & Pretrained $\pi_{0.5}$ \\
Fine-tuning & Full-parameter \\
Optimizer & AdamW \\
Learning rate & $5\times10^{-5}$ \\
Warmup steps & 100 \\
Global batch size & 128 \\
Training steps & 5,000 \\
Adam $(\beta_1,\beta_2)$ & $(0.9,\,0.95)$ \\
Weight decay & $10^{-10}$ \\
Gradient clipping norm & 1.0 \\
EMA decay & 0.999 \\
Action prediction horizon & 25 steps \\
Action history length & 1 step \\
Training prefix length & Uniform over $\{0,\ldots,9\}$ \\
\bottomrule
\end{tabular}
\end{table}

\noindent\textbf{Training configuration.}
Table~\ref{tab:training-config} summarizes the training hyperparameters.
We initialize the policy from pretrained $\pi_{0.5}$ weights and
fine-tune all parameters using AdamW. The learning rate reaches
$5\times10^{-5}$ after 100 warmup steps and remains constant thereafter.
Training uses a global batch size of 128, gradient clipping at a
global norm of 1.0, and an exponential moving average with decay 0.999.
The policy predicts 25 action steps per observation. For RTC training~\cite{rtc},
we randomly select 0 to 9 initial actions as a clean prefix,
which conditions the remaining predictions and is excluded from the
flow-matching loss. Per-step diffusion-time conditioning distinguishes
the clean prefix from the noisy actions.

\noindent\textbf{State and action representation.}
Each action concatenates the left-arm joint targets and gripper
command, the corresponding right-arm commands, and the base linear
and angular velocities, yielding a 16-dimensional vector. The state
uses the same ordering, combining the measured arm joint positions
and gripper states with the previous base velocity command.
The one-step history in Table~\ref{tab:training-config} supplies this
previous base command, providing motion context for the mobile base.
Historical actions are removed from the prediction targets, leaving
a 25-step action horizon. This history is separate from the RTC prefix,
which conditions predictions on actions retained from the preceding
action chunk.

\section{Extra Hardware Detials}

\begin{figure}[t]
    \centering
    \includegraphics[width=0.50\linewidth]{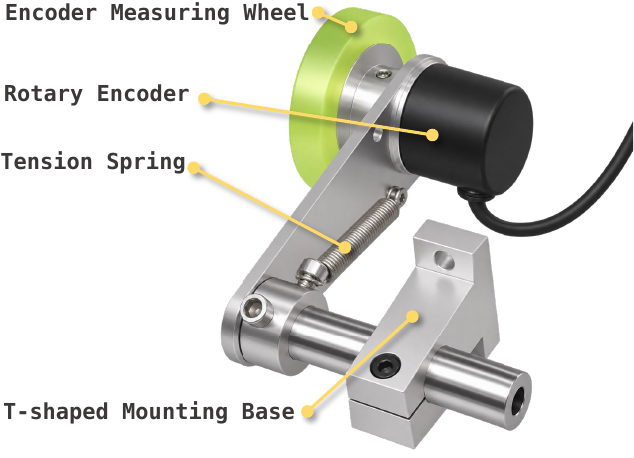}
    \caption{Off-the-shelf industrial measuring wheel assembly used for wheel odometry. The illustration was redrawn from a real photograph using GPT Image.}
    \label{fig:encoder}
\end{figure}

\noindent \textbf{Wheel odometry.}
Unlike Mobile ALOHA~\cite{mobile_aloha}, which requires a custom wheel odometry assembly with modified DYNAMIXEL motors and 3D-printed components, we adopt an off-the-shelf industrial measuring wheel assembly. As shown in~\cref{fig:encoder}, the assembly integrates an OMRON rotary encoder, a measuring wheel, and a spring-loaded mounting mechanism, eliminating the need for custom fabrication and motor modifications. We use a 300-pulse-per-revolution encoder, while higher-resolution variants are available, offering a resolution comparable to or exceeding that of the DYNAMIXEL encoder used in Mobile ALOHA.

\begin{figure}[t]
    \centering
    \includegraphics[width=0.75\linewidth]{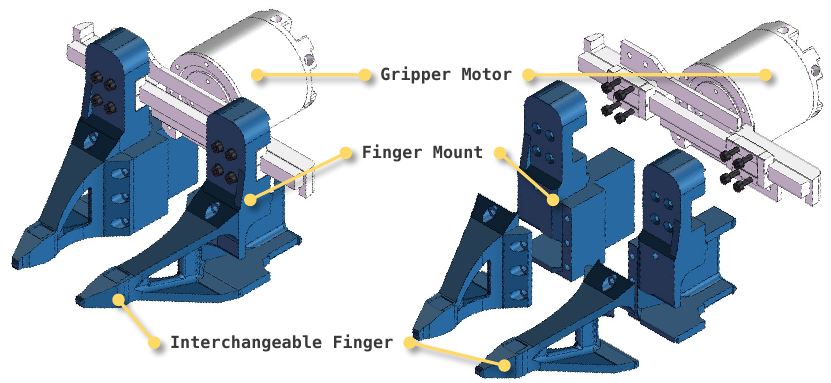}
    \caption{
        Modular end-effector design.
        Left: the assembled gripper.
        Right: an exploded view showing the gripper motor,
        finger mounts, and interchangeable fingers.
    }
    \label{fig:end_effector}
\end{figure}

\noindent \textbf{End effector desgin.}
As shown in Fig.~\ref{fig:end_effector}, in our released version, the end effector adopts
a modular design, two finger
mounts, and a pair of interchangeable fingers. The finger mounts
connect the fingers to the actuation unit, allowing the finger
geometry to be adapted to different manipulation tasks while
retaining the same actuation and mounting structure. The exploded
view illustrates the interfaces between these components.

\stopcontents[appendix]

\end{document}